\documentclass[]{spie}  

\usepackage{amsmath,amsfonts,amssymb}
\usepackage{graphicx}
\usepackage[colorlinks=true, allcolors=blue]{hyperref}

\title{SAM-DCE: Decoupled Context Enhancement for Prompt-Free Medical Image Segmentation}

\author[a]{Yingzhen Hu}
\author[b,c]{Yiheng Zhong}
\author[b,c]{Aijing Yang}
\author[c]{Georges El Fakhri}
\author[c]{Xiaofeng Liu}
\affil[a]{School of AI and Advanced Computing, Xi’an Jiaotong-Liverpool University, Suzhou, CN}
\affil[b]{Department of Biostatistics, Yale School of Public Health, Yale University, New Haven, CT 06519, USA}
\affil[c]{Biomedical Imaging Institute and Department of Radiology and Biomedical Imaging, Yale University, New Haven, CT, USA}

\begin{document} 
\maketitle

\begin{abstract}
The Segment Anything Model (SAM) is a promising foundation model for medical image segmentation, but its prompt-driven design is not directly suitable for fully automatic multi-class anatomical segmentation. Without manual point, box, or mask prompts, SAM faces a dual over-smoothing problem: deep image features may lose fine local structural evidence, while mask tokens may become semantically ambiguous. We propose SAM-DCE, a decoupled context enhancement framework for prompt-free medical image segmentation. SAM-DCE introduces Spatial Representation Compensation (SRC) to recover anatomical evidence from intermediate ViT features and Semantic Token Calibration (STC) to improve the category consistency of foreground mask tokens. Rather than replacing the pretrained decoder or fusing output masks, SAM-DCE refines the spatial and semantic representations before recombining them through SAM's original token-to-mask mechanism. On the Synapse multi-organ CT segmentation dataset, SAM-DCE achieved a mean Dice score of 83.86\% and an HD95 of 16.04 mm, outperforming the representative CNN-, Transformer-, and SAM-based baselines evaluated here. Ablation experiments further showed that SRC and STC provide complementary improvements, supporting the proposed spatial--semantic factorization for prompt-free SAM adaptation.
\end{abstract}


\section{Purpose}

The Segment Anything Model (SAM) has demonstrated strong general segmentation ability through large-scale pretraining on natural images~\cite{kirillov2023segment}, making it a promising foundation model for medical image segmentation~\cite{ma2024segment}. However, SAM is originally designed for prompt-driven interactive segmentation, where points, boxes, or masks specify the target object. In fully automatic medical image segmentation, manual prompts are usually unavailable, and the model must simultaneously locate anatomical structures and assign stable category identities.

Recent studies have adapted SAM to medical images through parameter-efficient tuning, prompt-free prediction, and medical-domain customization~\cite{zhang2023customized,shaharabany2023autosam,ma2024segment}. Nevertheless, prompt-free multi-class medical segmentation remains challenging because removing prompts exposes a dual over-smoothing problem. Spatially, repeated token mixing in the ViT encoder can weaken local anatomical evidence, especially for small organs, low-contrast tissues, and ambiguous boundaries~\cite{wang2022anti}. Semantically, SAM mask tokens are prompt-conditioned mask generators rather than fixed anatomical prototypes; without prompts, foreground tokens may receive similar contextual information and produce ambiguous category responses. This token-level degeneration is consistent with observations that deep ViT representations may lose discriminability while intermediate layers retain useful semantic diversity~\cite{ru2023token}.

These two failure modes arise in different representation spaces: spatial over-smoothing primarily affects image-derived mask features, whereas semantic over-smoothing affects mask tokens that must maintain stable category identities without prompts. This motivates addressing them separately rather than with a single undifferentiated decoder enhancement. We therefore propose SAM-DCE, a decoupled context enhancement framework for prompt-free medical image segmentation. Spatial Representation Compensation (SRC) recovers anatomical evidence from intermediate ViT features, while Semantic Token Calibration (STC) calibrates foreground mask tokens into more stable category queries. The compensated spatial representation and calibrated semantic tokens are recombined through SAM's original token-to-mask mechanism, preserving the pretrained decoding interface.

\section{Methods}

\begin{figure}[t]
\centering
\includegraphics[width=\linewidth]{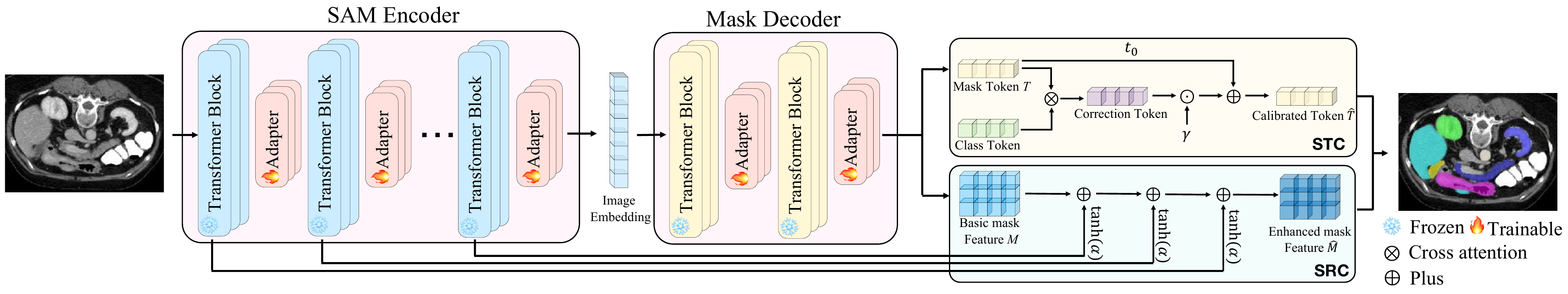}
\caption{
Overall framework of SAM-DCE. The model keeps the pretrained SAM backbone as the visual prior and introduces decoupled context enhancement in the decoder to compensate spatial representations and calibrate semantic tokens for prompt-free medical image segmentation.
}
\label{fig:framework}
\end{figure}

Given an input medical image $I$, SAM-DCE predicts a segmentation map with $C$ foreground anatomical categories and one background category without manual prompts. It builds on the pretrained SAM image encoder, prompt encoder, and two-way transformer, but enhances the internal decoder representations to address dual over-smoothing in prompt-free medical segmentation. The overall framework is shown in Fig.~\ref{fig:framework}.

The input image is encoded as $E=f_{\mathrm{enc}}(I)$, where $f_{\mathrm{enc}}(\cdot)$ is the pretrained SAM image encoder. Intermediate ViT features from the 3rd, 6th, 9th, and 12th Transformer blocks are retained as $\{F_3,F_6,F_9,F_{12}\}$. Since no prompts are used, the prompt encoder receives empty prompts and produces default prompt embeddings. The two-way transformer generates raw mask tokens $T=[t_0,t_1,\ldots,t_C]$, where $t_0$ is the background token and $t_1,\ldots,t_C$ are foreground tokens. Its decoded spatial feature is upscaled into the base mask feature $M$.

Let $\Phi(\cdot,\cdot)$ denote SAM's token-to-mask mapping. The raw prediction is $Y_{\mathrm{raw}}=\Phi(T,M)$. SAM-DCE constructs a spatially compensated mask feature $\hat{M}$ using SRC and calibrated mask tokens $\hat{T}$ using STC, giving $Y_{\mathrm{final}}=\Phi(\hat{T},\hat{M})$. Thus, rather than fusing independent branch predictions, SAM-DCE performs representation-level de-smoothing before mask generation: SRC restores the spatial evidence basis, while STC calibrates the semantic token basis.

\subsection{Decoupled Context Enhancement}

SRC compensates spatial over-smoothing by recovering local anatomical evidence from intermediate ViT features. For each $F_i$, $i\in\{3,6,9,12\}$, SRC projects it into a shared recovery space as $G_i=\mathcal{P}_i(F_i)$, where $\mathcal{P}_i(\cdot)$ denotes a lightweight projector with normalization, linear projection, and local convolutional refinement. Recovery starts from the deepest feature $G_{12}$ and progressively injects $G_9$, $G_6$, and $G_3$, establishing global semantic context first and then reintroducing increasingly local anatomical detail. Let $R_{12}$ be the initial recovered feature from $G_{12}$. For each shallower level $i\in\{9,6,3\}$, SRC updates the representation as $R_i=\mathcal{B}_i(\operatorname{Up}(R_j)+\tanh(\alpha_i)\operatorname{Up}(G_i))$, where $R_j$ is the previous deeper feature, $\operatorname{Up}(\cdot)$ is upsampling, $\mathcal{B}_i(\cdot)$ is a recovery block, and $\alpha_i$ is a learnable gate. The final recovered feature $R$ is added to the base mask feature as $\hat{M}=M+\tanh(\beta)\mathcal{P}_{\mathrm{out}}(R)$, where $\mathcal{P}_{\mathrm{out}}(\cdot)$ is the output projection and $\beta$ controls the residual strength. This residual formulation preserves the original SAM mask feature as the primary decoding pathway while allowing SRC to inject complementary anatomical detail. SRC also predicts a boundary map supervised by boundary labels extracted from the ground truth.

STC compensates semantic over-smoothing in the mask-token space. Given raw mask tokens $T=[t_0,t_1,\ldots,t_C]$, STC generates foreground semantic correction tokens $U=[u_1,\ldots,u_C]=\mathcal{C}(T)$, where $\mathcal{C}(\cdot)$ is the token calibration function. Each foreground token is updated as $\hat{t}_c=\operatorname{LN}(t_c+\sigma(\gamma)u_c)$ for $c=1,\ldots,C$, where $\operatorname{LN}(\cdot)$ is layer normalization, $\gamma$ is a learnable scaling parameter, and $\sigma(\cdot)$ constrains the calibration strength to $[0,1]$. The background token remains unchanged, restricting semantic calibration to foreground tokens whose anatomical category identities must be established without prompts, giving $\hat{T}=[t_0,\hat{t}_1,\ldots,\hat{t}_C]$.

\subsection{Training Objective}

Given the low-resolution ground-truth label $G$, SAM-DCE is trained with a main segmentation loss on $Y_{\mathrm{final}}$:
\begin{equation}
\mathcal{L}_{\mathrm{main}}
=
(1-\lambda)\mathcal{L}_{\mathrm{ce}}(Y_{\mathrm{final}},G)
+
\lambda\mathcal{L}_{\mathrm{dice}}(Y_{\mathrm{final}},G),
\end{equation}
where $\mathcal{L}_{\mathrm{ce}}$ and $\mathcal{L}_{\mathrm{dice}}$ are cross-entropy and Dice losses, and $\lambda$ balances them.

Two auxiliary predictions supervise the decoupled branches so that each corrected representation remains independently predictive before joint recombination: $Y_{\mathrm{stc}}=\Phi([t_0,U],M)$ for STC and $Y_{\mathrm{src}}=\Phi(T,\hat{M})$ for SRC. Both use the same cross-entropy and Dice formulation, producing $\mathcal{L}_{\mathrm{stc}}$ and $\mathcal{L}_{\mathrm{src}}$. The SRC boundary head is supervised by boundary labels $G_b$ obtained by identifying neighboring pixels with different class labels. The boundary objective combines binary cross-entropy and Dice-style overlap loss, producing $\mathcal{L}_{\mathrm{bd}}$.

The total objective is
\begin{equation}
\mathcal{L}
=
\mathcal{L}_{\mathrm{main}}
+
w_{\mathrm{stc}}\mathcal{L}_{\mathrm{stc}}
+
w_{\mathrm{src}}\mathcal{L}_{\mathrm{src}}
+
w_{\mathrm{bd}}\mathcal{L}_{\mathrm{bd}},
\end{equation}
where $w_{\mathrm{stc}}$, $w_{\mathrm{src}}$, and $w_{\mathrm{bd}}$ are loss weights. During training, the mask decoder and parameter-efficient adaptation parameters are optimized while retaining SAM's pretrained visual prior. During inference, SAM-DCE takes only the image as input and uses $Y_{\mathrm{final}}$ as the segmentation result.

\section{Results}

Table~\ref{tab:comparison-full} reports the quantitative comparison on the Synapse multi-organ segmentation dataset. Dice similarity coefficient (DSC) and 95th percentile Hausdorff distance (HD95) are used as the evaluation metrics. SAM-DCE achieves the best overall performance, with a mean Dice of \textbf{83.86\%} and an HD95 of \textbf{16.04 mm}. Compared with the SAM-based baseline SAMed, SAM-DCE improves the mean Dice from 80.26\% to 83.86\% and reduces HD95 from 28.89 mm to 16.04 mm. It achieves the best Dice scores on seven of eight organs, with particularly clear gains for the stomach and both kidneys. The gallbladder exception indicates that the benefit is not uniform across anatomical structures and that some small anatomical targets remain challenging. The improvements on structures requiring both fine localization and stable category assignment are consistent with the complementary roles of SRC and STC in prompt-free medical SAM adaptation.

\begin{table}[!t]
\caption{Performance comparison under the fully supervised setting.}
\label{tab:comparison-full}
\centering
\renewcommand{\arraystretch}{1.25}

\resizebox{\linewidth}{!}{%
\begin{tabular}{l|cccccccc|cc}
\hline

\textbf{Method}
& \textbf{Spleen}
& \textbf{Kidney(R)}
& \textbf{Kidney(L)}
& \textbf{Gallbladder}
& \textbf{Liver}
& \textbf{Stomach}
& \textbf{Aorta}
& \textbf{Pancreas}
& \textbf{Mean Dice[\%]}$\uparrow$
& \textbf{HD95[mm]}$\downarrow$ \\

\hline
\multicolumn{11}{c}{\textbf{Conventional Fully Supervised Methods}} \\
\hline

UNETR~\cite{hatamizadeh2022unetr}
& 85.47 & 81.60 & 78.82 & 58.15
& 93.78 & 65.68 & 82.96 & 44.01
& 73.81 & 69.49 \\

3D U-Net~\cite{cicek20163dunet}
& 88.41 & 80.94 & 80.30 & 53.55
& 95.33 & 72.19 & 87.41 & 52.46
& 76.32 & 39.21 \\

TransUNet~\cite{chen2021transunet}
& 87.23 & 63.13 & 81.87 & 77.02
& 94.08 & 55.86 & 85.08 & 75.62
& 77.48 & 31.69 \\

nn-UNet~\cite{isensee2021nnu}
& 89.52 & 81.97 & 81.38 & 57.47
& 95.41 & 66.59 & 90.90 & 65.21
& 78.56 & 55.94 \\

SwinUNet~\cite{cao2022swin}
& 85.47 & 66.53 & 83.28 & 79.61
& 94.29 & 56.58 & 90.66 & 76.60
& 79.13 & 21.55 \\

TransDeepLab~\cite{azad2022transdeeplab}
& 86.04 & 69.16 & 84.08 & 79.88
& 93.53 & 61.19 & 89.00 & 78.40
& 80.16 & 21.25 \\

\hline
\multicolumn{11}{c}{\textbf{SAM-based Methods}} \\
\hline

AutoSAM~\cite{shaharabany2023autosam}
& 80.54 & 80.02 & 79.66 & 41.37
& 89.24 & 61.14 & 82.56 & 44.22
& 62.08 & 27.56 \\

SAM Adapter~\cite{chen2023samadapter}
& 83.68 & 79.00 & 79.02 & 57.49
& 92.68 & 69.48 & 77.93 & 43.07
& 72.80 & 33.08 \\

SAMed~\cite{zhang2023customized}
& 87.33 & 80.10 & 82.75 & \textbf{70.24}
& 93.37 & 73.62 & 86.99 & 67.64
& 80.26 & 28.89 \\

\textbf{SAM-DCE (Ours)}
& \textbf{92.10}
& \textbf{84.90}
& \textbf{87.26}
& 66.94
& \textbf{95.13}
& \textbf{85.47}
& \textbf{88.28}
& \textbf{70.64}
& \textbf{83.86}
& \textbf{16.04} \\

\hline
\end{tabular}%
}

\end{table}

\begin{figure}[t]
\centering
\includegraphics[width=\linewidth]{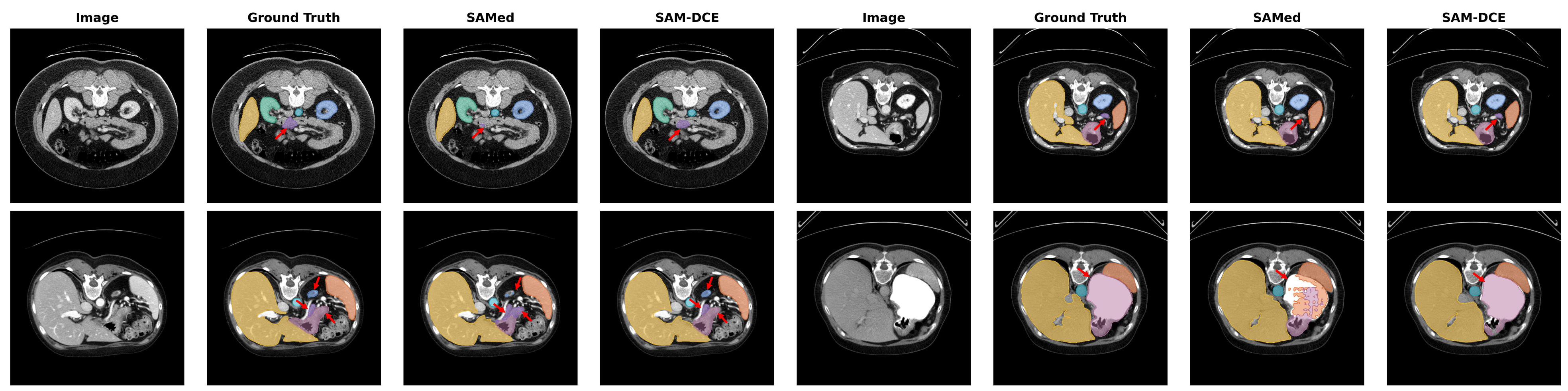}
\caption{
Qualitative comparison of SAMed and SAM-DCE on representative CT cases.
Red arrows indicate challenging regions characterized by small structures, low contrast, and ambiguous boundaries.
SAM-DCE provides more accurate anatomical delineation and better preserves local structural details compared with SAMed.
}
\label{fig:qualitative}
\end{figure}

Figure~\ref{fig:qualitative} shows representative qualitative comparisons between SAMed and SAM-DCE. Red arrows mark challenging regions with small structures, low contrast, or ambiguous boundaries. SAM-DCE produces more complete organ masks and cleaner anatomical contours, particularly in small or adjacent structures where SAMed tends to miss regions or produce less accurate boundaries. This behavior is consistent with SRC restoring local spatial evidence while STC stabilizes category-specific token responses.

Table~\ref{tab:src_stc_ablation} evaluates the contribution of SRC and STC. The baseline without both modules obtains 80.26\% Dice and 28.89 mm HD95. Adding STC improves Dice to 82.74\% and HD95 to 19.91 mm, while adding SRC improves Dice to 82.96\% and HD95 to 17.07 mm. Combining both modules achieves the best result, with 83.86\% Dice and 16.04 mm HD95, showing that spatial compensation and semantic token calibration are complementary. Among the individual modules, SRC yields the larger HD95 reduction, consistent with its explicit spatial recovery and boundary supervision, whereas STC still substantially improves both Dice and HD95. This is consistent with semantic calibration also contributing to spatial consistency through SAM's token-to-mask interaction rather than acting only on category identity. The further gain from combining SRC and STC supports the proposed factorization, indicating that the spatial and semantic corrections are complementary rather than redundant.

Table~\ref{tab:src_fusion} further compares different SRC fusion strategies. Single-level compensation yields the lowest HD95, highlighting the value of shallow features for precise boundary localization. Progressive recovery instead achieves the highest Dice, suggesting that gradually incorporating multiple feature depths improves overall region consistency by balancing high-level semantic context with local anatomical detail. Thus, the two metrics reveal a modest trade-off rather than uniform dominance of one fusion strategy.

A current limitation is that evaluation is restricted to a single abdominal CT benchmark. Future work will test generalization across modalities, anatomical targets, and SAM variants, and directly quantify spatial feature smoothness and token separability to further validate the proposed dual over-smoothing mechanism.

\begin{table}[!t]
\centering
\small

\begin{minipage}[t]{0.46\linewidth}
\vspace{0pt}
\centering
\caption{SRC and STC component ablation.}
\label{tab:src_stc_ablation}
\renewcommand{\arraystretch}{1.2}

\begin{tabular}{cc|cc}
\hline
\textbf{SRC}
& \textbf{STC}
& \textbf{Dice (\%)}$\uparrow$
& \textbf{HD95 (mm)}$\downarrow$ \\
\hline

$\times$ & $\times$
& 80.26 & 28.89 \\

$\times$ & $\checkmark$
& 82.74 & 19.91 \\

$\checkmark$ & $\times$
& 82.96 & 17.07 \\

$\checkmark$ & $\checkmark$
& \textbf{83.86} & \textbf{16.04} \\

\hline
\end{tabular}
\end{minipage}
\hfill
\begin{minipage}[t]{0.52\linewidth}
\vspace{0pt}
\centering
\caption{Ablation of SRC fusion strategies.}
\label{tab:src_fusion}
\renewcommand{\arraystretch}{1.2}

\begin{tabular}{l|cc}
\hline
\textbf{SRC Fusion Strategy}
& \textbf{Dice (\%)}$\uparrow$
& \textbf{HD95 (mm)}$\downarrow$ \\
\hline

Without SRC
& 80.26 & 28.89 \\

Single Block3
& 81.31 & \textbf{16.57} \\

Direct Multi-level Fusion
& 82.32 & 18.18 \\

Progressive Recovery
& \textbf{82.96} & 17.07 \\

\hline
\end{tabular}
\end{minipage}

\end{table}

 

\section{NOVEL OR BREAKTHROUGH ASPECTS}

$\bullet$ We formulate prompt-free medical SAM adaptation as a dual over-smoothing problem, involving both spatial degradation in image-derived mask features and semantic ambiguity in mask-token representations.

\noindent$\bullet$ We introduce a decoupled context enhancement design, where SRC recovers anatomical structure evidence from intermediate ViT features and STC calibrates foreground mask tokens into stable category queries.

\noindent$\bullet$ SAM-DCE performs factorized representation refinement before mask generation: SRC restores the spatial evidence basis and STC calibrates the semantic token basis, after which both are recombined through SAM's original token-to-mask mechanism without replacing the pretrained decoding interface.

\section{Conclusion}


SAM-DCE addresses prompt-free medical SAM adaptation by decoupling spatial representation recovery from semantic token calibration before mask generation. Experiments on Synapse show that the two components provide complementary improvements and together outperform existing SAM-based approaches. These results support spatial--semantic decoupling as a promising strategy for adapting prompt-driven foundation models to automatic multi-class medical segmentation.

\noindent \textbf{Disclosure of prior/concurrent submissions.} This work is not being, and has not been, submitted for publication or presentation elsewhere.

\bibliography{report} 
\bibliographystyle{spiebib} 

\end{document}